\documentclass[10pt,twocolumn,letterpaper]{article}

\usepackage{iccv}
\usepackage{times}
\usepackage{epsfig}
\usepackage{graphicx}
\usepackage{amsmath}
\usepackage{amssymb}
\usepackage{times}
\usepackage{epsfig}
\usepackage{graphicx}
\usepackage{amsmath}
\usepackage{amssymb}
\usepackage{color}

\usepackage[pagebackref=true,breaklinks=true,letterpaper=true,colorlinks,bookmarks=false]{hyperref}

\def\eg{\emph{e.g}. } 

\def\ie{\emph{i.e}. } 
\def\cf{\emph{c.f}. } 
\def\etc{\emph{etc}. } \def\vs{\emph{vs}. }
 
\def\etal{\emph{et al}. }

\iccvfinalcopy 

\def\iccvPaperID{28} 

\ificcvfinal\fi
\begin{document}

\title{Drought Stress Classification using 3D Plant Models}

\author{Siddharth Srivastava\thanks{Equal Contribution}\hspace{2mm}\textsuperscript{\textdagger}, Swati Bhugra \footnotemark[1]\hspace{2mm}\textsuperscript{\textdagger}, Brejesh Lall\textsuperscript{\textdagger}, Santanu Chaudhury\textsuperscript{\textdagger}\textsuperscript{\textdaggerdbl} \\
\textsuperscript{\textdagger}Department of Electrical Engineering, Indian Institute of Technology, Delhi, India\\
\textsuperscript{\textdaggerdbl}Central Electronics Engineering Research Institute, Pilani, India\\
{\tt\small \{eez127506, eez138301, brejesh, santanuc\}@ee.iitd.ac.in}
}

	\maketitle
	
\def\etal{et al. }
\def\etc{etc. }
\def\ie{i.e. }
\def\eg{e.g. }
\def\cf{cf. }
\def\vs{vs. }
\def\L{\mathcal{L}}
\def\T{\mathcal{T}}
\def\x{\textbf{x}}
\def\R{\mathbb{R}}

\begin{abstract}
Quantification of physiological changes in plants can capture different drought mechanisms and assist in selection of tolerant varieties in a high throughput manner.  In this context, an accurate 3D model of plant canopy provides a reliable representation for drought stress characterization in contrast to using 2D images. In this paper, we propose a novel end-to-end pipeline including 3D reconstruction, segmentation and feature extraction, leveraging deep neural networks at various stages, for drought stress study. To overcome the high degree of self-similarities and self-occlusions in plant canopy, prior knowledge of leaf shape based on features from deep siamese network are used to construct an accurate 3D model using structure from motion on wheat plants. The drought stress is characterized with a deep network based feature aggregation. We compare the proposed methodology on several descriptors, and show that the network outperforms conventional methods.
\end{abstract}

\section{Introduction}

Drought stress is a primary factor for limiting crop productivity \cite{humplik2015automated}. Thus, there is an urgent need for breeding high yielding cultivars. Quantification of physiological traits (plant phenotyping) can explain diverse drought stress responses and assist in selection of these cultivars in a high throughput manner. Current methods for drought stress study are predominantly based on extraction of features from 2D images \cite{humplik2015automated}. Due to high self-occlusions in plants, this results in information loss since 2D images are canopy projections on a plane \cite{cai2013smart}. 
Thus, precise 3D modeling of plant canopy is required for an accurate quantification of different phenotypic traits such as wilting, biomass etc. for drought stress analysis. 

Recently, many methods that use image samples to directly model the plant are primarily based on visual hull \cite{kutulakos1999theory} or Multi-View Stereo (MVS) \cite{seitz2006comparison}. Kumar et al. \cite{kumar2014high} employed  visual hull algorithm for 3D reconstruction with a static plant canopy and camera rotating at fixed height around it using a turntable setup. Cai and Miklavcic \cite{cai2012automated} utilized 2D skeletons to overcome the difficulties such as overlapping plant parts and broken segments for a smooth 3D reconstruction. Kumar et al. \cite{kumar2012high} presented a mirror based setup that enabled the back of the plant to be captured in the front view, however, it resulted in loss of resolution. Visual hull methods for reconstruction of thin leaf surfaces with discontinuities in plant canopy often result in overestimated models. 
In contrast to the previously mentioned approaches, authors in \cite{rose2015accuracy} employed multi-view stereo and Structure from Motion (SfM) to obtain initial sparse point cloud  and then patch based MVS (PMVS) was used to obtain dense point clouds to represent basil, tomato plants  and mint leaves. Lou et al. \cite{lou2014accurate} also utilized SfM followed by stereo matching and depth-map merging process for 3D plant modeling. These studies are suitable for plants with broad leaves but for thin leaved plants (For Ex: wheat, rice) such approaches generate point cloud with hole and gaps. Thus, Pound et al. \cite{pound2014automated} presented a patch based method to obtain dense model of rice and wheat canopy. They utilized correspondence based methods \cite{furukawa2010accurate,wu2011visualsfm} to obtain initial point clouds and these points are segmented into small patches (leaf segments) developed individually using level sets, which optimizes the model based on neighboring information. But, the cluster size to obtain leaf level segmentation before the level set step depends on the complexity of the plant structure and is a user driven parameter. \par 

The methods discussed previously indicate that the 3D modeling of plants is a challenging task due to high self-occlusions and leaves spanning arbitrary directions \cite{weber1995creation}. Thus, in contrast  to the aforementioned image based methods, we propose to use a prior on the leaf shape of the plant for a robust and dense 3D reconstruction. This is followed by 3D segmentation to obtain the point cloud corresponding only to the plant canopy from the reconstruction. Another challenge is the extraction of relevant traits that successfully encapsulates the physiological changes in response to drought. Therefore, motivated by recent success of deep neural networks, in this work we propose to take a learning based approach over 3D models of plants for computing such traits. The 3D features are directly extracted from the obtained point cloud using a deep neural network and is fused with learned local features from the same model for aggregating local and global information. We compare the features from the deep neural network with several baselines based on deep and local 3D descriptors to demonstrate the effectiveness of the learned features in characterizing the subtle differences in the plant 
architecture under drought stress. In view of the above, the following are the contributions of this paper:
 
\begin{itemize}
\item We reconstruct a 3D model of plants at mature growth stages with high degree of occluded leaves, by infusing keypoints and descriptors from a deep network to a Structure from Motion (SfM) pipeline. We show that this method is better than traditional keypoint detector and descriptors used with SfM for 3D plant reconstruction.
\item We propose and evaluate an aggregation of learned global and local features with a deep neural network for 3D Point Cloud. We show that the learned features are capable of encoding structural and visual changes in plant during drought stress. Although many methods have been proposed for the purpose of drought stress identification \cite{singh2016machine,bhugra2017phenotyping}, but to the best of our knowledge, this is the first work that utilizes deep networks on 3D data to learn an implicit representation of features for drought stress characterization. We evaluate the features from deep networks against conventional descriptors, The reliability of different 3D features are shown based on drought stress classification demonstrating that deep networks provide better accuracy and lesser computational complexity at test time.    
\end{itemize}

The rest of the paper is organized as follows. In Section \ref{sec:relatedwork}, we discuss literature related to the proposed work followed by methodology in Section \ref{sec:approach}. Finally, Section \ref{sec:conclusion} contains the concluding comments.

\section{Related Work}\label{sec:relatedwork}

Lang \cite{lang1973leaf} presented a contact based method by employing a mechanical arm with potentiometers to touch the plant surface and record its joint rotations to obtain 3D models. This method is semi-invasive as the apparatus can change the target canopy structure. Another approach \cite{sinoquet1991estimating} used sonic digitizer, where the pointer was an ultrasound emitter and based on the time intervals between emission and reception of sound in the sensor, the position was computed. But, this approach is sensitive to the structure of the plant canopy and wind. These contact based techniques are labor intensive and low throughput as experts are needed to register the measurement.
Another set of approaches is to create plant models based on compact user defined rules. Generative L-systems \cite{quan2006image} rules motivated by plant growth and relational growth models have been applied to a variety of problems. These methods are used for creating synthetic plant structures but they do not capture the detailed structure of real plants and the parameters used for their synthesis are difficult to use for a non-expert. \par
Light Detection and Ranging (LiDAR) sensing technology has also been utilized to obtain 3D models of plants.  Authors in \cite{kaminuma2004automatic} employed laser scanners to model 3D surface of leaves and petioles as polygonal meshes of \textit{Arabidopsis thaliana}. Li et al. \cite{rakocevic2000assessing} presented a framework to track and detect plant growth by a forward-backward 3D point cloud analysis, where the 3D point cloud was produced based on a structured light scanner over time. Mesh based 3D LIDAR approach was proposed by Paproki et al. \cite{paproki2012novel} where the plant was partitioned into morphological regions. The robustness of the presented method was based on the calculation of Leaf Area Index (LAI). Sirault et al. \cite{sirault2013plantscan} fused PMVS, voxel coloring and LiDAR data using registration algorithms in their digitizing platform. The camera was calibrated using a fixed camera setup with potted plants on a precise turntable. Although these methods can deal with complex plant boundaries, some laser-based approaches fail in direct sunlight. Moreover, the scanning time increases with the resolution of the point cloud and it requires expensive equipment inaccessible to many.  \par
In contrast to LiDAR approaches, Kinect sensor systems simultaneously capture both depth and color images thus, making it suitable for phenotypic analysis. Azzari et al. \cite{azzari2013rapid} utilized a Microsoft Kinect sensor combined with the point cloud library to obtain the depth images and extract proxy indices for plant volume. Alternatively, Cai  \cite{cai2014integration} integrated  both visible image and Kinect depth map to compute a robust depth estimate. Since the Kinect sensor has a comparably low-resolution, the depth estimation at object boundaries becomes unreliable. Thus, it may not be able to capture the 3D information of plants with narrow leaves. The narrow leaves can fall between the key points of the emitted pattern, resulting in decoding errors of the structured light patterns. 
 
Recently, several architectures have been proposed with deep neural networks for plant phenotyping \cite{pound2016deep, namin2017deep, deepplant} and leaf segmentation \cite{romera2015recurrent,ren2016end}. While they do not address the problem of drought stress, these work aim at 2D imaging modality instead of 3D. Moreover, deep neural networks capable of processing 3D data have come up very recently \cite{maturana2015voxnet, riegler2016octnet, Qi_2017_CVPR}, explaining the lack of representative work on plant phenotyping with 3D and deep neural networks.
\section{Methodology}\label{sec:approach}

\begin{figure*}	
	\centering
	\includegraphics[scale=0.495]{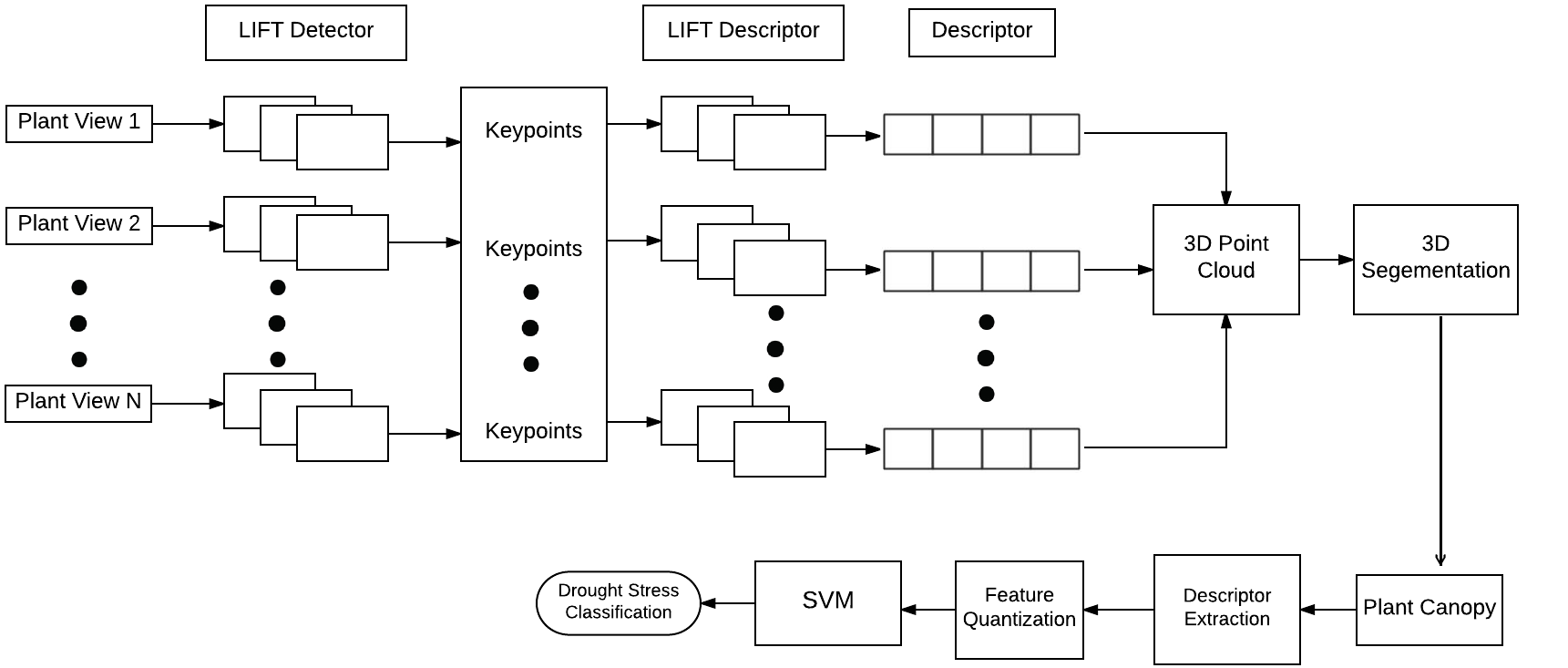}
	\caption{Flow diagram for the proposed approach}
	\label{fig:overall}
\end{figure*}

The end to end pipeline is shown in Figure \ref{fig:overall}. The proposed approach begins with 3D reconstruction using Structure from Motion where we use learned keypoints and descriptors obtained by fine-tuning a deep network. This is followed by segmenting the 3D plant canopy from the reconstructed point cloud. After this, techniques based on local and deep descriptors are used to directly extract relevant features from the point cloud. These features are then utilized for drought stress classification of wheat plants.

\subsection{3D Plant Reconstruction using Learned Invariant Feature Transform}

Our 3D reconstruction pipeline utilizes recent progress in deep networks \cite{maturana2015voxnet, riegler2016octnet, Qi_2017_CVPR}. Specifically, we modify the standard SfM pipeline as proposed in \cite{snavely2006photo} to use learned keypoints and descriptors based on deep networks. We employ Learned Invariant Feature Transform (LIFT) \cite{yi2016lift} for learning the keypoint detector and descriptor instead of the Scale-Invariant Feature Transform (SIFT) \cite{lowe1999object}. The output of the proposed framework is shown in Figure \ref{fig:mv}. The authors in \cite{snavely2006photo} used SIFT detector and descriptors for finding correspondence among images. SIFT is a hand-crafted feature where keypoints are found by scale-space analysis to identify the most discriminative and transform invariant regions in an image while the descriptor encapsulates the information within a pre-defined region around the detected keypoints. On the other hand, keypoints in LIFT correspond to  distinctive regions, where the conditions defining distinctiveness are learned with a deep siamese network on the dataset for the target domain. Additionally, the SIFT detector and descriptor are designed to work independently of each other, while in case of LIFT, the learning of detector and descriptor is achieved with the help of an end-to-end pipeline. Therefore, SIFT is suitable for applications where the characteristics of an image follow the underlying assumptions behind the design of SIFT while LIFT aims at adapting to the distinctive characteristics within the images of the problem under consideration.  

In our initial experiments, we observed that SIFT resulted in point clouds with holes along the base and leaf tips. The reason being that images of a plant from different viewing angles appeared similar and suffered from heavy occlusion of leaves in wheat plants (Figure \ref{fig:mv}), resulting in false matches by SIFT in a few regions. We performed experiments with Speeded-Up Robust features (SURF) \cite{bay2006surf} as well, and observed the same problems (holes). Since the structure of plants is complicated (thin, smooth, heavily occluded, highly similar leaves), we resorted to the current setting of leveraging learned keypoints and descriptors for assisting the reconstruction pipeline. As discussed above, due to the self similar, low texture regions and high degree of occlusion in plant images, SIFT and SURF failed to characterize regions (lack of sufficient and appropriate keypoints) for acceptable point cloud reconstructions. Many recent works utilize deep networks to learn patch based descriptors \cite{simo2015discriminative, zagoruyko2015learning,yi2016lift,han2015matchnet} from images. Therefore, we selected LIFT \cite{yi2016lift} for its ability to learn keypoints and corresponding descriptor based upon specific characteristics of the dataset. This becomes important in the current scenario since plant images present unique challenges unlike other type of images. 

Learning deep network based patch descriptors and detectors requires us to train the detector-descriptor with specific examples. This allows the network to adjust to intricacies in the structure of the plants, especially the curvature and variation in color (a key component for identifying drought stress) etc. We perform training of the LIFT detector-orientation estimator-descriptor pipeline with a \textit{leaf correspondence dataset} (described ahead) following the methodology proposed by the authors of LIFT. The \textit{leaf correspondence dataset} consists of $10,210$ images of various plants captured from various viewing angles with different cameras. We use these images, to construct 3D models using VisualSFM \cite{wu2011visualsfm} with SIFT features. In total, we reconstruct $90$ 3D models from approximately $90$-$110$ views per plant. After this, positive and negative samples are formed based on whether the respective keypoints are preserved or not, respectively, in the corresponding 3D reconstruction using SfM. We then extract the training patches following the methodology suggested by the authors of LIFT. The  patches thus extracted provide additional information during the training process as the used regions from the 3D model are robust from multiple views and hence the details from the surrounding regions can be exploited by the deep siamese network. 

Authors in \cite{schonberger2017comparative} provide a comparative evaluation of various descriptors including those based on deep networks on various image related tasks. They found that a few variations of SIFT such as SIFTRoot, SIFTPCA performed better in a structure from motion pipeline than learned descriptors. However, the datasets on which SfM pipelines are usually evaluated have rigid structures, while the current use case involves plants where the objects are non-rigid, thin and highly similar. Moreover, the images are captured in a green house with no control on lighting and movement of objects in the surrounding with variations in the position of the leaves (due to air etc.) posing additional challenges. Therefore, it becomes important in our case to have significant number of correspondences on the plant itself, unlike earlier techniques on plant 3D reconstruction  where background information was used as an indicator for camera parameter estimation and subsequently 3D reconstruction \cite{pound2014automated}.

\subsection{3D Segmentation using Voxel Cloud Connectivity Segmentation}

We use Voxel Cloud Connectivity Segmentation (VCCS) \cite{papon2013voxel} for segmenting the plant canopy from the reconstructed point cloud. It works directly on 3D point clouds. While segmenting leaves in 2D images is an active area of research \cite{CVPP2015_4, CVPP2015_5}, due to availability of depth information the problem in 3D can be looked at from a pure computer vision perspective. Here a plant's relative placement with respect to the surroundings can be leveraged. Moreover, in our case the background is significantly distinctive than the plant itself. Since VCCS utilizes spatial location ($x$,$y$,$z$) as well RGB information associated with points to perform segmentation, we directly use it for segmenting the point cloud.   
The method consists of converting the input point cloud to a voxelized point cloud and building an adjacency graph. For constructing the adjacency graph, a voxel grid is formed and seed voxels are selected and initialized. The isolated voxels are filtered by considering a small search volume around the seed voxels. Voxels are then clustered, conditioned upon the smallest gradient within the search volume. The clusters are further aggregated into supervoxels by comparing a $39$ dimensional feature vector derived using XYZ, RGB etc. from respective cluster centers.  The aggregation proceeds in a breadth first manner to provide the final segmentation. 

\subsection{3D Features: Local and Deep Descriptors}

Our next step in the pipeline is to extract 3D features from the point cloud. We consider two types of descriptors (i) Local Descriptors (ii) Deep Descriptors. The former includes the class of descriptors which rely on finding keypoint in 3D point clouds and then describing it with the help of its neighborhood. The latter involves using deep neural networks for learning features from 3D point clouds. We now describe each of these in the following subsections.
 \begin{figure*}[ht]
	\centering
	\includegraphics[scale=0.52]{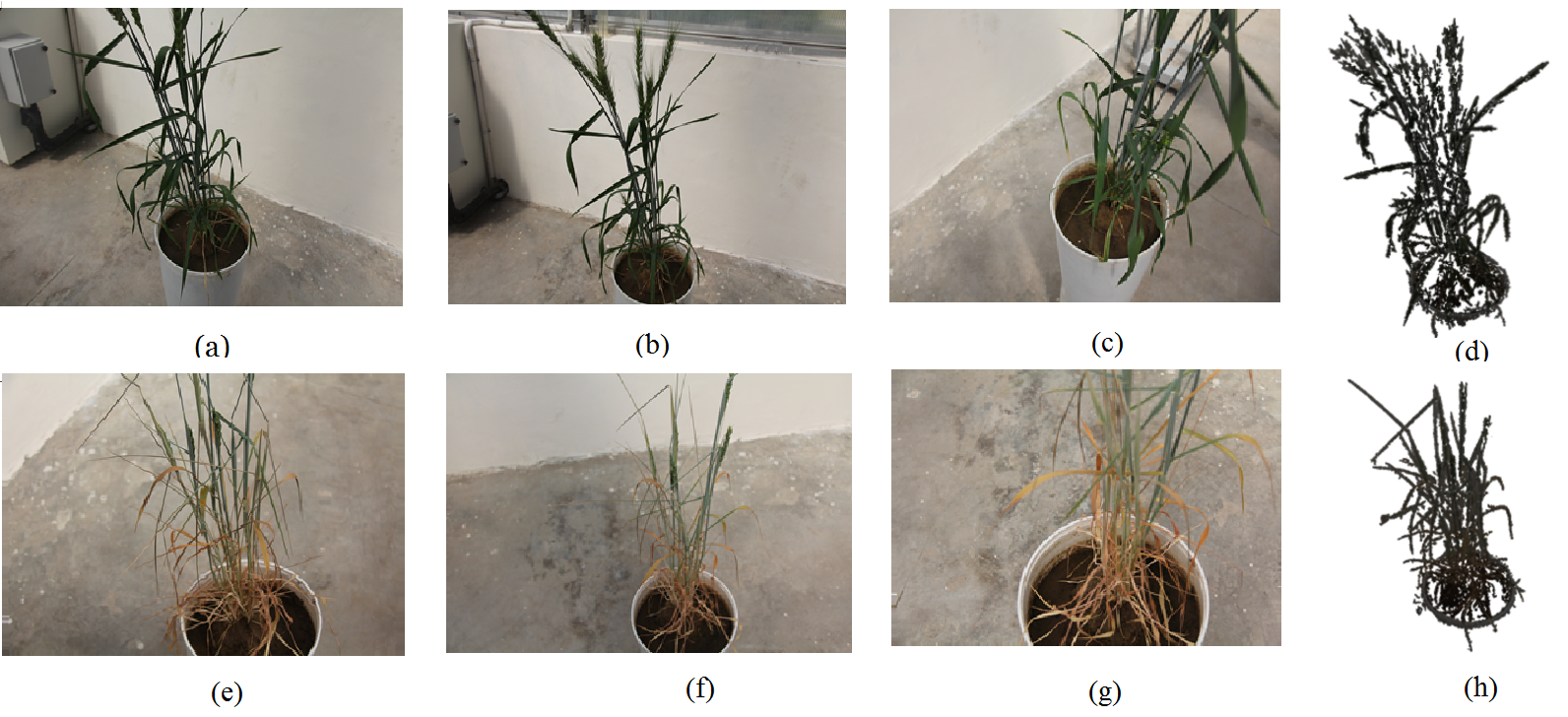}
	\caption{Sample views of a wheat plant under control [(a)-(c)] and drought stress [(e)-(g)] condition in the phenomics facility. The last column [(d),(h)]  shows the corresponding 3D reconstruction using LIFT keypoints and descriptors (models are shown without RGB rendering for clarity). The figure is shown for one drought plant, the dataset consists of plants with varying drought stress levels.}
	\label{fig:mv}
\end{figure*}

\subsubsection{Local Descriptors}

We evaluate Signature of Histograms of Oriented Gradients (SHOT) \cite{salti2014shot}, Rotational Projection Statistics (RoPS) \cite{Guo2013} and Fast Point Feature Histograms (FPFH) \cite{rusu2009fast} for drought stress classification. With comprehensive analysis, Guo et al. \cite{guo2016comprehensive} showed that these descriptors provide superior results on a  variety of benchmark tasks involving 3D point clouds. However, applicability and comparison of these descriptors in characterizing various plant related tasks, and specifically drought stress identification, has not been studied earlier. 

The primary advantage with local descriptors is in their ability to encode geometrical properties of the model. This characteristic makes them suitable for quantifying various phenotypic traits involving structural changes. We attempt to leverage this characteristic of local descriptors for drought stress identification, where the leaves undergo various structural changes depending upon the amount of stress.

\subsubsection{Deep Descriptors}

There are two types of deep architectures to process 3D data: (i) 3D Convolutional Neural Networks (3D-CNN) \cite{maturana2015voxnet, qi2016volumetric} (ii) PointNet \cite{Qi_2017_CVPR}. Due to the inherent nature of the convolution operation, the 3D-CNNs work on structured data, i.e., voxelized cloud. However, PointNet is a recent architecture that works directly on unstructured 3D Point Cloud data. Voxelization of the point cloud introduces approximation to the model as it is essentially a quantization process. Therefore, we adopt PointNet as the deep network for further processing. 

PointNet generates a global feature on the input point cloud. This is done by learning a permutation invariant representation of the points from the input point cloud which is encoded into a vector using a symmetric function. The invariance to transformation is achieved using a joint alignment network which essentially predicts an affine transformation matrix and applies it to the point cloud while features from multiple point clouds are aligned using a feature transformation matrix. However, by design, the global feature produced by PointNet does not capture local geometric information. We observed this to be the reason for relatively poor performance of global features from PointNet trained on traditional objects as discussed in Section \ref{sec:experiments}. In order to overcome this limitation, we aggregate the local and global information similar to PointNet's segmentation network, i.e., the global descriptor is fed back to the network along with the descriptor of the keypoint to generate a more robust keypoint descriptor. Next, we quantize the local descriptors thus obtained for the keypoints detected on the point cloud and concatenate it with the the global descriptor. The motivation being that such a fusion would make the resultant (global) descriptor encode both local and global information. Here, local information encodes fine changes in the surface and color of a plant while experiencing the drought, while the global information encapsulates the overall change in the structure of the plant, possibly such as leaf rolling, color variations over multiple leaves, wilting etc. As will be shown in experiments, the aggregation of local information provides significant performance gains.

\subsection{Drought Stress Classification}

In order to classify the objects, we follow the classification pipeline shown in Figure \ref{fig:overall}. The pipeline begins by extracting 3D features from segmented 3D point cloud (Plant Canopy). This is followed by a training and testing phase. It must be noted that the number of keypoints (and hence the local descriptors) are different for each point cloud. Therefore, during training phase, we learn a quantized feature representation of the 3D features. Quantization is necessary to obtain a single descriptor of uniform length for each point cloud. We experimented with both Fisher Vector \cite{sanchez2013image} and Bag of Visual Words \cite{csurka2004visual} and found that Fisher Vector works better in our case. Our training set consists of 3D point cloud of wheat plants (note it is not the same set of images on which LIFT was trained). In literature, Fisher Vector has mostly been used with SIFT features \cite{gosselin2014revisiting}, while it has not been found suitable directly for depth data \cite{bo2011depth}. However  authors in \cite{cheng2015convolutional} propose a Convolutional Fisher Kernel including preprocessing steps which allows Fisher Vector to efficiently encode depth data as well. In this work, we directly quantize the features using Fisher Vector. As will be evident in the experimental section, the Fisher Vectors are able to discriminatively encode the point cloud features. Since the size of the codebook from Fisher Vector is small, we use a linear classifier, Support Vector Machine (SVM) for further classification. This setting allows us to reduce the overall computational complexity while maintaining robustness.
\section{Experimental Results}\label{sec:experiments}

\subsection{Dataset} 
\textit{Data Collection}: The drought experiment was conducted on wheat pots at the Plant Phenomics Facility, Indian Agricultural Research Institute (IARI), Pusa, New-Delhi during Rabi season of 2016-17. Two replicates of each genotype of wheat plants were studied. For each pair, one was grown in well-watered conditions while the other was subjected to water deficit conditions for a period of 5 continuous days. The images were taken by manually moving the visible camera (Canon 60D EOS) with eight mega-pixel resolution around the plant. A few sample images of the plants are shown in Figure \ref{fig:mv}.  \\
\\
\textit{Dataset Details}: For experiments reported in the current study, the dataset consists of $3,200$ images having a resolution of $5184$ x $3456$ pixels each for $34$ wheat plants. Out of these, $17$ plants belongs to the control category while the rest, under water-deficit conditions, belong to the drought stress category. For each plant, we took $80$-$100$ images from various angles and distances in an indoor environment with varying background depending upon the size and complexity of occlusions in a plant. The training set consists of $2304$ images from $24$ healthy and drought stress plants while test set comprises of the rest of the images.


\subsection{Results}

\begin{table*}[]
\centering
\caption{Classification Accuracy in 3D and Feature Computation Time}
\label{tab:tab1}
\begin{tabular}{|l|r|r|}
\hline
\multicolumn{1}{|c|}{\textbf{Descriptor}} & \multicolumn{1}{c|}{\textbf{\begin{tabular}[c]{@{}c@{}}Accuracy \\ (\%)\end{tabular}}} & \multicolumn{1}{c|}{\textbf{\begin{tabular}[c]{@{}c@{}}Computation Time (sec)\\ {[}Average Per Model{]}\end{tabular}}} \\ \hline
SHOT (FV)                                 & 76.0                                                                                   & 5.3                                                                                                                    \\ \hline
SHOT (BoVW)                               & 74.2                                                                                   & 6.8                                                                                                                    \\ \hline
RoPS (FV)                                 & 77.2                                                                                   & 4.9                                                                                                                    \\ \hline
RoPS (BoVW)                               & 75.4                                                                                   & 5.2                                                                                                                    \\ \hline
FPFH (FV)                                 & 73.3                                                                                   & 3.9                                                                                                                    \\ \hline
FPFH (BoVW)                               & 72.1                                                                                   & 4.3                                                                                                                    \\ \hline
PointNet (Global)                         & 65.4                                                                                   & \textbf{2.3}                                                                                                           \\ \hline
PointNet (Aggregation)                    & 67.2                                                                                   & 2.36                                                                                                                   \\ \hline
Fine tuned PointNet (Global)              & 76.3                                                                                   & 2.4                                                                                                                    \\ \hline
Fine tuned PointNet (Aggregation)         & \textbf{79.1}                                                                          & 2.5                                                                                                                    \\ \hline
\end{tabular}
\end{table*}

\textit{Baseline}: Due to lack of prior studies on performance of quantization technique with 3D descriptors, especially in case of plants, we report results on both Fisher Vector (FV) and Bag of Visual Words (BoVW), which are amongst the most popular feature quantization techniques. We use the common dataset as described above for training and testing various methods. For PointNet, we report results on both pre-trained model and after fine-tuning PointNet with 3D Point Cloud models from the training dataset. Here, the pre-trained model refers to the PointNet trained for the task of object classification \cite{Qi_2017_CVPR}. The fine-tuning is performed by initializing the weights from the pre-trained PointNet for object classification and then continuing the training process with the 3D point cloud of the wheat plants. Further, the results are reported for both global descriptor (\textit{PointNet(global)} and \textit{Fine tuned PointNet (global)}) and aggregated descriptor (\textit{PointNet (aggregation)} and \textit{fine tuned PointNet (aggregration)}), for both pre-trained and fine-tuned PointNet respectively. 

\textit{Qualitative Results}: Figure \ref{fig:mv} (column $4$) shows a view of the reconstructed 3D model of a control and drought plant. One can observe that the base of reconstructed model of the drought plant have clean reconstruction despite heavy occlusion. Similar observation can be made for control plant where leaves are occluded throughout the plant structure, while also having smooth curvature in some leaves. Reconstruction of such fine details can be attributed to the quality of matches on the leaf surfaces from learned keypoints and descriptors for 3D reconstruction. This also shows potential for the technique to be generalized to other types of plants as well.

\textit{Quantitative Results}: We report comparative evaluation on classification accuracy and computational complexity of the proposed methodology. The accuracy is computed as the percentage of number of correct classifications to the total number of test inputs for respective classes i.e. drought and healthy. Table \ref{tab:tab1} shows that the fine-tuned PointNet with aggregation descriptor outperforms all the other techniques with the closest being RoPS (FV) by $1.9$\%, followed by fine-tuned PointNet (Global) by $2.8$\%. However, it is interesting to note that pre-trained PointNet on rigid objects performs poorly against all the compared descriptors. This could be due to two reasons (i) the default architecture of PointNet is not easily generalizable, and, (ii) as discussed earlier, plants have smooth and textureless surface and are usually heavily occluded, which are not usually found in rigid bodies. The good performance of fine-tuned PointNet further strengthens the argument that the proposed aggregration of features is indeed able to characterize the structural and visual changes such as wilting, color variations etc. in the plant.  

Further, RoPS (FV) outperforms all the other local descriptors followed by SHOT (FV). It can be seen that descriptors quantized with Fisher Vector consistently perform better than the corresponding encoding with Bag of Visual Words with gap on an average being $1.6$\%. This shows that as with 2D descriptors, Fisher Vector is able to encode a more discriminative representation of local descriptors as compared to BoVW. Therefore, in the aggregration of feaures in fine-tuned PointNet, Fisher Vector was used as the feature quantization technique.

The computation time shown in Table \ref{tab:tab1} is computed by summing the average description time for each keypoint along with quantization (excluding PointNet) and classification for a model averaged over all the 3D models in the test dataset. It can be seen that deep descriptors are nearly twice as fast as local descriptors at test time.  However, we do note that while it takes significantly less time for computing descriptor and classifying a point cloud at test time, it took approximately $4$x more time than the local descriptors to train the network.

\section{Conclusion}\label{sec:conclusion}

We proposed a novel end-to-end automated pipeline for drought stress classification in plants in 3D. We performed exhaustive experiments and demonstrated the effectiveness of the proposed methodology on wheat plants. We showed that deep descriptors fine tuned on plant point clouds perform better than local descriptors. However, we also showed that deep descriptors on point clouds without fine tuning perform worse than local descriptors, which mandates the need to pursue efforts in this direction for publicly available large datasets of plants. In future works, the proposed work can also be used for analyzing characteristic changes in plant architecture in response to other abiotic stresses.
	
{\small
\bibliographystyle{ieee}
\bibliography{IEEEfull}

\begin{thebibliography}{10}\itemsep=-1pt

\bibitem{azzari2013rapid}
G.~Azzari, M.~L. Goulden, and R.~B. Rusu.
\newblock Rapid characterization of vegetation structure with a microsoft
  kinect sensor.
\newblock {\em Sensors}, 13(2):2384--2398, 2013.

\bibitem{bay2006surf}
H.~Bay, T.~Tuytelaars, and L.~Van~Gool.
\newblock Surf: Speeded up robust features.
\newblock {\em Computer vision--ECCV 2006}, pages 404--417, 2006.

\bibitem{bhugra2017phenotyping}
S.~Bhugra, A.~Anupama, S.~Chaudhury, B.~Lall, and A.~Chugh.
\newblock Phenotyping of xylem vessels for drought stress analysis in rice.
\newblock In {\em Machine Vision Applications (MVA), 2017 Fifteenth IAPR
  International Conference on}, pages 428--431. IEEE, 2017.

\bibitem{bo2011depth}
L.~Bo, X.~Ren, and D.~Fox.
\newblock Depth kernel descriptors for object recognition.
\newblock In {\em Intelligent Robots and Systems (IROS), 2011 IEEE/RSJ
  International Conference on}, pages 821--826. IEEE, 2011.

\bibitem{cai2014integration}
J.~Cai.
\newblock The integration of images and kinect depth maps for better quality of
  3d surface reconstruction.
\newblock In {\em Control Automation Robotics \& Vision (ICARCV), 2014 13th
  International Conference on}, pages 223--227. IEEE, 2014.

\bibitem{cai2012automated}
J.~Cai and S.~Miklavcic.
\newblock Automated extraction of three-dimensional cereal plant structures
  from two-dimensional orthographic images.
\newblock {\em IET Image Processing}, 6(6):687--696, 2012.

\bibitem{cai2013smart}
X.~Cai, Y.~Sun, Y.~Zhao, L.~Damerow, P.~S. Lammers, W.~Sun, J.~Lin, L.~Zheng,
  and Y.~Tang.
\newblock Smart detection of leaf wilting by 3d image processing and 2d fourier
  transform.
\newblock {\em Computers and electronics in agriculture}, 90:68--75, 2013.

\bibitem{cheng2015convolutional}
Y.~Cheng, R.~Cai, X.~Zhao, and K.~Huang.
\newblock Convolutional fisher kernels for rgb-d object recognition.
\newblock In {\em 3D Vision (3DV), 2015 International Conference on}, pages
  135--143. IEEE, 2015.

\bibitem{csurka2004visual}
G.~Csurka, C.~Dance, L.~Fan, J.~Willamowski, and C.~Bray.
\newblock Visual categorization with bags of keypoints.
\newblock In {\em Workshop on statistical learning in computer vision, ECCV},
  volume~1, pages 1--2. Prague, 2004.

\bibitem{CVPP2015_5}
M.~Dyrmann.
\newblock Fuzzy c-means based plant segmentation with distance dependent
  threshold.
\newblock In H.~S. S.~A.~Tsaftaris and T.~Pridmore, editors, {\em Proceedings
  of the Computer Vision Problems in Plant Phenotyping (CVPPP)}, pages
  5.1--5.11. BMVA Press, September 2015.

\bibitem{furukawa2010accurate}
Y.~Furukawa and J.~Ponce.
\newblock Accurate, dense, and robust multiview stereopsis.
\newblock {\em IEEE transactions on pattern analysis and machine intelligence},
  32(8):1362--1376, 2010.

\bibitem{gosselin2014revisiting}
P.-H. Gosselin, N.~Murray, H.~J{\'e}gou, and F.~Perronnin.
\newblock Revisiting the fisher vector for fine-grained classification.
\newblock {\em Pattern Recognition Letters}, 49:92--98, 2014.

\bibitem{guo2016comprehensive}
Y.~Guo, M.~Bennamoun, F.~Sohel, M.~Lu, J.~Wan, and N.~M. Kwok.
\newblock A comprehensive performance evaluation of 3d local feature
  descriptors.
\newblock {\em International Journal of Computer Vision}, 116(1):66, 2016.

\bibitem{Guo2013}
Y.~Guo, F.~Sohel, M.~Bennamoun, M.~Lu, and J.~Wan.
\newblock Rotational projection statistics for 3d local surface description and
  object recognition.
\newblock {\em International Journal of Computer Vision}, 105(1):63--86, Oct
  2013.

\bibitem{han2015matchnet}
X.~Han, T.~Leung, Y.~Jia, R.~Sukthankar, and A.~C. Berg.
\newblock Matchnet: Unifying feature and metric learning for patch-based
  matching.
\newblock In {\em Proceedings of the IEEE Conference on Computer Vision and
  Pattern Recognition}, pages 3279--3286, 2015.

\bibitem{humplik2015automated}
J.~F. Humpl{\'\i}k, D.~Laz{\'a}r, A.~Husi{\v{c}}kov{\'a}, and L.~Sp{\'\i}chal.
\newblock Automated phenotyping of plant shoots using imaging methods for
  analysis of plant stress responses--a review.
\newblock {\em Plant methods}, 11(1):29, 2015.

\bibitem{kaminuma2004automatic}
E.~Kaminuma, N.~Heida, Y.~Tsumoto, N.~Yamamoto, N.~Goto, N.~Okamoto,
  A.~Konagaya, M.~Matsui, and T.~Toyoda.
\newblock Automatic quantification of morphological traits via
  three-dimensional measurement of arabidopsis.
\newblock {\em The Plant Journal}, 38(2):358--365, 2004.

\bibitem{kumar2012high}
P.~Kumar, J.~Cai, and S.~Miklavcic.
\newblock High-throughput 3d modelling of plants for phenotypic analysis.
\newblock In {\em Proceedings of the 27th conference on image and vision
  computing New Zealand}, pages 301--306. ACM, 2012.

\bibitem{kumar2014high}
P.~Kumar, J.~Connor, and S.~Mikiavcic.
\newblock High-throughput 3d reconstruction of plant shoots for phenotyping.
\newblock In {\em Control Automation Robotics \& Vision (ICARCV), 2014 13th
  International Conference on}, pages 211--216. IEEE, 2014.

\bibitem{kutulakos1999theory}
K.~N. Kutulakos and S.~M. Seitz.
\newblock A theory of shape by space carving.
\newblock In {\em Computer Vision, 1999. The Proceedings of the Seventh IEEE
  International Conference on}, volume~1, pages 307--314. IEEE, 1999.

\bibitem{lang1973leaf}
A.~Lang.
\newblock Leaf orientation of a cotton plant.
\newblock {\em Agricultural Meteorology}, 11:37--51, 1973.

\bibitem{lou2014accurate}
L.~Lou, Y.~Liu, J.~Han, and J.~H. Doonan.
\newblock Accurate multi-view stereo 3d reconstruction for cost-effective plant
  phenotyping.
\newblock In {\em International Conference Image Analysis and Recognition},
  pages 349--356. Springer, 2014.

\bibitem{lowe1999object}
D.~G. Lowe.
\newblock Object recognition from local scale-invariant features.
\newblock In {\em Computer vision, 1999. The proceedings of the seventh IEEE
  international conference on}, volume~2, pages 1150--1157. Ieee, 1999.

\bibitem{maturana2015voxnet}
D.~Maturana and S.~Scherer.
\newblock Voxnet: A 3d convolutional neural network for real-time object
  recognition.
\newblock In {\em Intelligent Robots and Systems (IROS), 2015 IEEE/RSJ
  International Conference on}, pages 922--928. IEEE, 2015.

\bibitem{namin2017deep}
S.~T. Namin, M.~Esmaeilzadeh, M.~Najafi, T.~B. Brown, and J.~O. Borevitz.
\newblock Deep phenotyping: Deep learning for temporal phenotype/genotype
  classification.
\newblock {\em bioRxiv}, page 134205, 2017.

\bibitem{papon2013voxel}
J.~Papon, A.~Abramov, M.~Schoeler, and F.~Worgotter.
\newblock Voxel cloud connectivity segmentation-supervoxels for point clouds.
\newblock In {\em Proceedings of the IEEE Conference on Computer Vision and
  Pattern Recognition}, pages 2027--2034, 2013.

\bibitem{paproki2012novel}
A.~Paproki, X.~Sirault, S.~Berry, R.~Furbank, and J.~Fripp.
\newblock A novel mesh processing based technique for 3d plant analysis.
\newblock {\em BMC Plant Biology}, 12(1):63, 2012.

\bibitem{pound2016deep}
M.~P. Pound, A.~J. Burgess, M.~H. Wilson, J.~A. Atkinson, M.~Griffiths, A.~S.
  Jackson, A.~Bulat, G.~Tzimiropoulos, D.~M. Wells, E.~H. Murchie, et~al.
\newblock Deep machine learning provides state-of-the-art performance in
  image-based plant phenotyping.
\newblock {\em bioRxiv}, page 053033, 2016.

\bibitem{pound2014automated}
M.~P. Pound, A.~P. French, E.~H. Murchie, and T.~P. Pridmore.
\newblock Automated recovery of three-dimensional models of plant shoots from
  multiple color images.
\newblock {\em Plant Physiology}, 166(4):1688--1698, 2014.

\bibitem{Qi_2017_CVPR}
C.~R. Qi, H.~Su, K.~Mo, and L.~J. Guibas.
\newblock Pointnet: Deep learning on point sets for 3d classification and
  segmentation.
\newblock In {\em The IEEE Conference on Computer Vision and Pattern
  Recognition (CVPR)}, July 2017.

\bibitem{qi2016volumetric}
C.~R. Qi, H.~Su, M.~Nie{\ss}ner, A.~Dai, M.~Yan, and L.~J. Guibas.
\newblock Volumetric and multi-view cnns for object classification on 3d data.
\newblock In {\em Proceedings of the IEEE Conference on Computer Vision and
  Pattern Recognition}, pages 5648--5656, 2016.

\bibitem{quan2006image}
L.~Quan, P.~Tan, G.~Zeng, L.~Yuan, J.~Wang, and S.~B. Kang.
\newblock Image-based plant modeling.
\newblock In {\em ACM Transactions on Graphics (TOG)}, volume~25, pages
  599--604. ACM, 2006.

\bibitem{rakocevic2000assessing}
M.~Rakocevic, H.~Sinoquet, A.~Christophe, and C.~Varlet-Grancher.
\newblock Assessing the geometric structure of a white clover (trifolium repens
  l.) canopy using3-d digitising.
\newblock {\em Annals of Botany}, 86(3):519--526, 2000.

\bibitem{ren2016end}
M.~Ren and R.~S. Zemel.
\newblock End-to-end instance segmentation and counting with recurrent
  attention.
\newblock {\em CVPR 2017}, 2016.

\bibitem{riegler2016octnet}
G.~Riegler, A.~O. Ulusoys, and A.~Geiger.
\newblock Octnet: Learning deep 3d representations at high resolutions.
\newblock {\em CVPR}, 2016.

\bibitem{romera2015recurrent}
B.~Romera-Paredes and P.~H. Torr.
\newblock Recurrent instance segmentation.
\newblock {\em CVPR 2017}, 2015.

\bibitem{rose2015accuracy}
J.~C. Rose, S.~Paulus, and H.~Kuhlmann.
\newblock Accuracy analysis of a multi-view stereo approach for phenotyping of
  tomato plants at the organ level.
\newblock {\em Sensors}, 15(5):9651--9665, 2015.

\bibitem{rusu2009fast}
R.~B. Rusu, N.~Blodow, and M.~Beetz.
\newblock Fast point feature histograms (fpfh) for 3d registration.
\newblock In {\em Robotics and Automation, 2009. ICRA'09. IEEE International
  Conference on}, pages 3212--3217. IEEE, 2009.

\bibitem{salti2014shot}
S.~Salti, F.~Tombari, and L.~Di~Stefano.
\newblock Shot: Unique signatures of histograms for surface and texture
  description.
\newblock {\em Computer Vision and Image Understanding}, 125:251--264, 2014.

\bibitem{sanchez2013image}
J.~S{\'a}nchez, F.~Perronnin, T.~Mensink, and J.~Verbeek.
\newblock Image classification with the fisher vector: Theory and practice.
\newblock {\em International journal of computer vision}, 105(3):222--245,
  2013.

\bibitem{schonberger2017comparative}
J.~L. Sch{\"o}nberger, H.~Hardmeier, T.~Sattler, and M.~Pollefeys.
\newblock Comparative evaluation of hand-crafted and learned local features.
\newblock In {\em Conference on Computer Vision and Pattern Recognition
  (CVPR)}, 2017.

\bibitem{seitz2006comparison}
S.~M. Seitz, B.~Curless, J.~Diebel, D.~Scharstein, and R.~Szeliski.
\newblock A comparison and evaluation of multi-view stereo reconstruction
  algorithms.
\newblock In {\em Computer vision and pattern recognition, 2006 IEEE Computer
  Society Conference on}, volume~1, pages 519--528. IEEE, 2006.

\bibitem{CVPP2015_4}
K.~Simek and K.~Barnard.
\newblock Gaussian process shape models for bayesian segmentation of plant
  leaves.
\newblock In H.~S. S.~A.~Tsaftaris and T.~Pridmore, editors, {\em Proceedings
  of the Computer Vision Problems in Plant Phenotyping (CVPPP)}, pages
  4.1--4.11. BMVA Press, September 2015.

\bibitem{simo2015discriminative}
E.~Simo-Serra, E.~Trulls, L.~Ferraz, I.~Kokkinos, P.~Fua, and F.~Moreno-Noguer.
\newblock Discriminative learning of deep convolutional feature point
  descriptors.
\newblock In {\em Proceedings of the IEEE International Conference on Computer
  Vision}, pages 118--126, 2015.

\bibitem{singh2016machine}
A.~Singh, B.~Ganapathysubramanian, A.~K. Singh, and S.~Sarkar.
\newblock Machine learning for high-throughput stress phenotyping in plants.
\newblock {\em Trends in plant science}, 21(2):110--124, 2016.

\bibitem{sinoquet1991estimating}
H.~Sinoquet, B.~Moulia, and R.~Bonhomme.
\newblock Estimating the three-dimensional geometry of a maize crop as an input
  of radiation models: comparison between three-dimensional digitizing and
  plant profiles.
\newblock {\em Agricultural and Forest Meteorology}, 55(3-4):233--249, 1991.

\bibitem{sirault2013plantscan}
X.~R.~R. Sirault, J.~Fripp, A.~Paproki, P.~Kuffner, C.~Nguyen, R.~Li, H.~Daily,
  J.~Guo, and R.~Furbank.
\newblock Plantscan™: a three-dimensional phenotyping platform for capturing
  the structural dynamic of plant development and growth.
\newblock {\em FSPM2013 Proceedings}, 2013.

\bibitem{snavely2006photo}
N.~Snavely, S.~M. Seitz, and R.~Szeliski.
\newblock Photo tourism: exploring photo collections in 3d.
\newblock In {\em ACM transactions on graphics (TOG)}, volume~25, pages
  835--846. ACM, 2006.

\bibitem{deepplant}
J.~R. Ubbens and I.~K. Stavness.
\newblock Ddeep plant phenomics: A deep learning platform for complex plant
  phenotyping tasks.
\newblock {\em Frontiers in Plant Science}, 2017.

\bibitem{weber1995creation}
J.~Weber and J.~Penn.
\newblock Creation and rendering of realistic trees.
\newblock In {\em Proceedings of the 22nd annual conference on Computer
  graphics and interactive techniques}, pages 119--128. ACM, 1995.

\bibitem{wu2011visualsfm}
C.~Wu et~al.
\newblock Visualsfm: A visual structure from motion system.
\newblock 2011.

\bibitem{yi2016lift}
K.~M. Yi, E.~Trulls, V.~Lepetit, and P.~Fua.
\newblock Lift: Learned invariant feature transform.
\newblock In {\em European Conference on Computer Vision}, pages 467--483.
  Springer, 2016.

\bibitem{zagoruyko2015learning}
S.~Zagoruyko and N.~Komodakis.
\newblock Learning to compare image patches via convolutional neural networks.
\newblock In {\em Proceedings of the IEEE Conference on Computer Vision and
  Pattern Recognition}, pages 4353--4361, 2015.

\end{thebibliography}
}

\end{document}